\documentclass{article}

\usepackage[nonatbib,preprint]{neurips_2022}
\usepackage[numbers]{natbib}
\usepackage{mathtools}

\usepackage[utf8]{inputenc}
\usepackage[T1]{fontenc}
\usepackage{hyperref}
\usepackage{url}
\usepackage{booktabs}
\usepackage{amsfonts}
\usepackage{nicefrac}
\usepackage{microtype}
\usepackage{xcolor}

\let\originalparagraph\paragraph
\renewcommand{\paragraph}[2][.]{\originalparagraph{#2#1}}

\title{Redundancy in Deep Linear Neural Networks}
\author{ Oriel BenShmuel 
\\
Faculty of Math\&CS \\
Weizmann Institute of Science \\
Israel \\
\texttt{oriel.benshmuel@weizmann.ac.il}
}
\date{\vspace{-5ex}}

\begin{document}

\date{\vspace{-5ex}}

\maketitle

\begin{abstract}
Conventional wisdom states that deep linear neural networks benefit from expressiveness and optimization advantages over a single linear layer. This paper suggests that, in practice, the training process of deep linear fully-connected networks using conventional optimizers is convex in the same manner as a single linear fully-connected layer. This paper aims to explain this claim and demonstrate it. Even though convolutional networks are not aligned with this description, this work aims to attain a new conceptual understanding of fully-connected linear networks that might shed light on the possible constraints of convolutional settings and non-linear architectures.
\end{abstract}

\section{Introduction}
Linear networks provide a simple and elegant solution for elementary problems such as simple classification \cite{fan2008liblinear,zhang2001text} and regression problems \cite{weisberg2005applied,hanin2021data}. Moreover, linear models are suitable for approximate higher-level representation of the dataset distribution using AutoEncoders \cite{bourlard1988auto,multilayerAE,kodirov2017semantic,meng2017relational}. They could serve as classic denoising solvers \cite{pretorius2018learning} or even datasets' manifold estimators \cite{hinton1997modeling}. However, linear models perform poorly on complex problems expressed with nonlinear properties \cite{murphy2006comparative,lu2017expressive,raghu2017expressive}.

On the other hand, deep neural networks with nonlinear abilities are an extensively researched field. These nonlinear models reached state-of-the-art performance on computer vision problems \cite{voulodimos2018deep,redmon2016you,he2017mask,ren2015faster,goodfellow2014generative} and natural language processing tasks \cite{vaswani2017attention,devlin2018bert,bahdanau2014neural,brown2020language}. Yet, the clarity of the learning process and the direct influence of the samples on the network remained quite vague. Therefore, deep linear networks provide a convenient framework to understand part of the bigger picture of deep learning.

Recent findings in theoretical deep-learning show that despite the linear mapping between the input and the output of deep linear networks, they still have optimization advantages over a single linear layer network \cite{arora2018optimization,hardt2018identity,kawaguchi2016deep,saxe2014exact}. These findings contain an elaborated explanation for this claim using theoretical analysis and experimental demonstrations.

According to the theoretical analysis, deep linear networks have a non-convex optimization process. This paper demonstrates how this perception might be different for fully-connected linear networks. Deep fully-connected linear networks could perform a non-convex (and non-concave) optimization process. However, in practice, these networks are, indeed, going through a convex optimization process that is experimentally equivalent to a single fully-connected linear layer network.

We begin in section \ref{propvow} by exposing some properties of fully connected linear networks trained with stochastic gradient descent (SGD \cite{spall2005introduction}) and experimentally support our claims in section \ref{exp}. Then, in section \ref{opt}, we show how these properties lead to an equivalent optimization process of a deep linear network and a single linear layer (with fully-connected architectures). Finally, in section \ref{mom}, we explain why SGD with momentum \cite{pmlr-v28-sutskever13} also performs an equivalent optimization process.

\section{Proportions in fully-connected linear networks}
\label{propvow}
This section will expose some properties regarding the weights of fully-connected linear networks (experimentally supported by section \ref{exp}). We will use the following notations:

\begin{enumerate}
    \item $\theta_l$ of size $k_l\times n_l$ - The weights of layer $l$ in the network.
    \item $\theta_l[j]$ - The $j$th row of $\theta_l$.
    \item $\partial \theta$ - The changes of the weights in the first step with respect to the gradient of the loss function.
\end{enumerate}

For a network with a single linear layer, we will define $k_1=2$ (for two classes). Without loss of generality, we will focus on the randomly initialized weights $\theta_1[0]$ (and not $\theta_1[1]$) as the vector of weights related to the first output neuron. For an input space of size $n$, the size of vector $\theta_1[0]$ is $1\times n$. 

\begin{paragraph}{Claim 1}
\label{claim:samp1layer1}
Let $x$ be a single training sample used for training a network with a single linear layer for a single step. Then there is a scalar $\alpha\in\mathbb{R}$ such that:
$$\partial \theta_1[0]=\alpha x$$
\end{paragraph}

\begin{paragraph}{Claim 2}
\label{claim:sampnlayer1}
Given a network with a single linear layer with randomly initialized weights $\theta_1$ and a set $\{(x_i,\alpha_i)\}_{i=1}^b$ such that each pair corresponds to the proportional property described in \emph{Claim 1} with respect to $\theta_1$. Training the network with the entire batch $\{x_i\}_{i=1}^b$ for a single step (with the same initial weights $\theta_1$) will result in the following equality:
$$\partial \theta_1[0]= \frac{1}{b}\sum_{i=1}^b{\alpha_i x_i}$$
\end{paragraph}

\begin{paragraph}{Claim 3}
\label{claim:sampnlayern}
For a deep linear network, the following statement is applied (using the previous notations):
$$\forall_{0\leq j<k_1}\exists{r_j\in \mathbb{R}}: \partial \theta_1[j]= r_j\frac{1}{b}\sum_{i=1}^b{\alpha_i x_i}$$
\end{paragraph}

\paragraph{Corollary 1}
For a deep linear network:
$$\forall_{0\leq j_1,j_2<k_1}\exists r_{j_1},r_{j_2}\in \mathbb{R}: r_{j_2}\cdot\partial \theta_1[j_1]=r_{j_1}\cdot\partial \theta_1[j_2]$$

\begin{paragraph}{Claim 4}
\label{claim:evol}
For a deep linear network, we get the following for any layer $l$ in the network and any step in the training process:
$$\forall_{0<l<depth, 0\leq j_1,j_2<k_l} \exists{r\in \mathbb{R}}: \partial \theta_l[j_1]=r\cdot\partial \theta_l[j_2]$$
\end{paragraph}

\section{Experimental support}
\label{exp}
To measure how close the vectors are, in terms of proportionally, for each angle $a>90^{\circ}$ in the experiments, we use $180^{\circ}-a$ instead. In addition, we are using the negative log likelihood loss function (a non-linear function). The experiments were conducted with GPU K80.

\paragraph{Claim 1 (support)}
For a single linear layer, we will use the classes \emph{Cat} and \emph{Dog} of the dataset CIFAR10 \cite{krizhevsky2009learning}. For ten different initializations of a single linear layer, we randomly pick $100$ different samples. We compute the angle between $\partial \theta_1[0]$ and the chosen sample. The mean value of the calculated angles is $0.01^{\circ}$, and the standard deviation is $0.005^{\circ}$.

As expected, each angle is very close to $0^{\circ}$ or to $180^{\circ}$ (up to numerical errors), which indicates that $\partial \theta_1[0]=\alpha x$, for some sample $x$ and scalar $\alpha\in \mathbb{R}$.

\paragraph{Claim 2 (support)}
For a batch size of $30$ and a single linear network with a single layer, we get that the expression $||\partial \theta_1[0]-\frac{1}{b}\sum_{i=1}^b{\alpha_i x_i}||$ over ten initializations produces an average value of $3.62\cdot 10^{-8}$. It implies that the equation $\partial \theta_1[0]= \frac{1}{b}\sum_{i=1}^b{\alpha_i x_i}$ is indeed true up to numerical errors.

\paragraph{Claim 3 (support)}
In the same manner, for multiple layers in the network (using ``Architecture B'' appears in Appendix \ref{appendix:architectures}) and a batch size of $30$, consider all possible combinations of $\partial \theta_1[j_1]$ and $\partial \theta_1[j_2]$ for $0\leq j_1\neq j2<k_1$. We get that for the angles above $90^{\circ}$, we have an average angle of $179.99^{\circ}$, and for the angles below $90^{\circ}$, the average angle is $0.001^{\circ}$ which supports the fact that: $$\forall_{0\leq j_1,j_2<k_1}\exists r\in \mathbb{R}:\partial \theta_1[j_1]=r\cdot\partial \theta_1[j_2]$$

\paragraph{Claim 4 (support)}
For long-term training of linear network with multiple layers, the following experiments will include the average angle between the vectors $\{\partial \theta_l^t[j]\}_{0\leq j <k_l}$ for each layer $l$ as a function of step number $t$. Each analysis will also include a graph of the model's accuracy over the training steps. 

We will use several linear architectures for the experiments. The architectures appear in Appendix \ref{appendix:architectures}.

\begin{figure}[htp!]
     \includegraphics[width=1\textwidth]{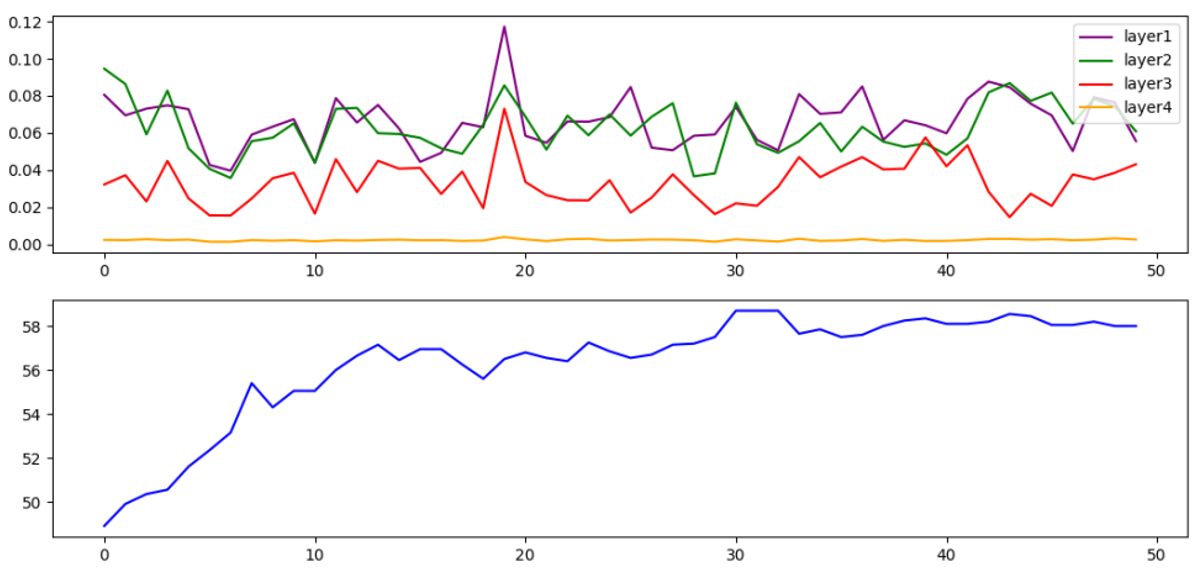}
     \caption{The top graph describes the average angle between $\partial \theta_l^t[j_1]$ and $\partial \theta_l^t[j_2]$ for each layer $l$ as a function of steps $t$. The bottom graph describes the model's accuracy.}
     \label{fig:app_redundancy_3vs5_50}
\end{figure}

\noindent
In Figure \ref{fig:app_redundancy_3vs5_50}, we can see the analysis for \emph{Cat} versus \emph{Dog} trained with a batch size of $128$, a learning rate of $1e-2$, and ``Architecture B'' over $50$ steps.
The first image shows the average angle between each pair of vectors ($\partial \theta_l^t[j_1], \partial \theta_l^t[j_2]$ for $0\leq j_1\neq j_2<k_l$) as a function of the iteration $t$ (the step number). There are four plots in the first image, each plot for each one of the four linear layers in the network. The presented angles are in \emph{degrees}, and it is easy to spot that (up to a complement of $180^{\circ}$) the angle is below a single degree (which is extremely small). The second graph (below the first one) shows the accuracy of the same network. It implies that the angles are independent of the accuracy, and up to minor errors, they have very small values in each phase of the training. Overall it supports the claim that for any given step $t$ and layer $l$, the following expression is true:
$$\forall_{0\leq j_1,j_2<k_l} \exists_{r\in \mathbb{R}}: \partial \theta_l^t[j_1]\approx r\cdot\partial \theta_l^t[j_2]$$
\noindent
An additional set of experiments with multiple different settings appears in Appendix \ref{appendix:additional_results_linear}.

\noindent
Note: When calculating the angles, we normalize the vectors. For vectors with small norms in the first place, a numerical error might occur during the normalization. Therefore, wider layers might have more significant errors. Overall the error is tiny (the vertical scale of the first plot is in degrees) and usually below $0.5^{\circ}$.

\section{The optimization process}
\label{opt}
Following the above claims (section \ref{propvow}), for each layer $l$, the update of the weights $\partial \theta_l$ could be observed as a collection of the vectors $\{t_j\cdot \partial \theta_l[0]\}_{0\leq j<k}$ (for some set of scalars $t_j \in \mathbb{R} $) rather than a collection of the vectors $\{\partial  \theta_l[j]\}_{0\leq j<k}$. In this case, the entire matrix $\partial \theta_l$ depends on a single vector $\theta_l[0]$ up to scalar multiplication. Additionally, in the classic training process, where we use stochastic gradient steps (SGD), this is true for any given step in the training process (and not only for the first step). In other words, we get that each layer is updated with a weights matrix of rank $1$. Their multiplication does not reduce the expressiveness of the network.

\begin{figure}[htp]
    \centering
     \includegraphics[width=0.7\textwidth]{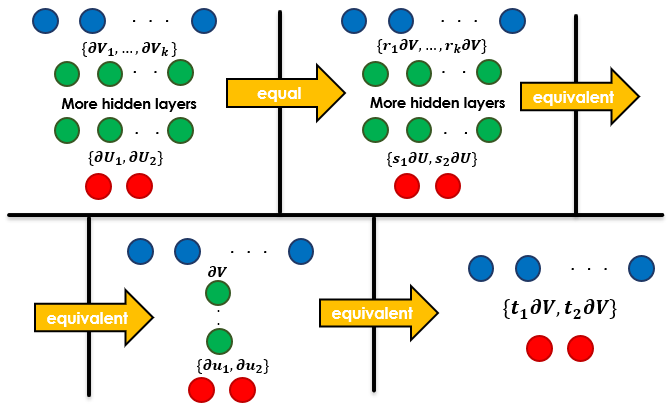}
     \caption{Visualization of the weights' update's equivalency transitions from a deep linear network to a single linear layer.}
     \label{fig:reduction}
\end{figure}

In general, we can apply a simple reduction from the optimization process of fully-connected deep linear networks to the optimization process of a single fully-connected layer, as illustrated in Figure \ref{fig:reduction}. Assuming that:
$$\forall_{0\leq j_1,j_2<k_l,0<l<depth} \exists_{r\in \mathbb{R}}: \partial \theta_l[j_1]=r\cdot\partial \theta_l[j_2]$$
we can represent $\partial \theta_l$ as the collection of the vectors $\{t_j\cdot \partial \theta_l[0]\}_{0\leq j<k_l}$. Since $\partial \theta_l$ depends on a single vector, up to scalar multiplication, the same update in the weights could have been done with a single neuron (which also depends on a single vector of weights of the same size). Extending this conclusion for each step and layer $l$ in the original network (excluding the output layer), we get a similar optimization process of a fully-connected linear network with only a single neuron for each hidden layer. In the case of a single neuron in each hidden layer, we get a network that is not using its depth since the weights of each hidden layer is a scalar. Therefore, the network could be represented as a single linear layer in the training process as an equivalent learner to any deep fully-connected linear architecture.

This section concludes that the training process of a randomly initialized fully-connected deep linear network is experimentally equivalent to a randomly initialized linear network with a single layer.

\section{SGD with Momentum}
\label{mom}
In many cases, the SGD optimizer is used with momentum \cite{pmlr-v28-sutskever13} to acquire convergence advantages. For such an optimizer, the optimization process enjoys memorization abilities based on the weights update of previous steps. The momentum algorithm is conducted as follows:
$$\mathcal{V}^{t+1}=\beta\cdot \mathcal{V}^{t}+(1-\beta)\cdot \partial \theta^t$$
$$\theta^{t+1}=\theta^t-\alpha \cdot \mathcal{V}^{t+1}$$
Where $t$ is the iteration index, $\alpha$ is the learning rate, and $\beta$ is the momentum factor.
For an iterative representation of $\mathcal{V}^n$, we get:
$$\mathcal{V}^n=(1-\beta)\cdot \sum^{n}_{t=1}{\beta^{n-t}\partial \theta^t}$$

Assume a new optimization method with a predefined number of steps (in our case, $n$ steps), using the traditional SGD with a gentle twist:
$$\gamma_t = (1-\beta)\cdot \sum^{n}_{i=t}{\beta^{i-t}}$$

$$\theta^{t+1}=\theta^t-\alpha \cdot \gamma_t \cdot \theta^t$$

In the new variant, $\theta^t$ is an iteration-dependent scalar. Proportional vectors would be proportional even after scalar multiplications. Therefore, the properties mentioned in section \ref{propvow} and supported by section \ref{exp} are applied to the new variant of SGD defined above. Moreover, we get the following equality:
$$\sum^{n}_{s=1}{\mathcal{V}^s}=\sum^{n}_{s=1}{\sum^{s}_{t=1}{(1-\beta)\beta^{s-t}\partial \theta^t}}=$$
$$(1-\beta)\bigg(\partial \theta^1(\beta^0+...+\beta^{n-1})+\partial \theta^2(\beta^0+...+\beta^{n-2})+...+\theta^{n}(\beta^0)\bigg)=$$
$$=(1-\beta)\bigg(\partial \theta^1\sum^{n-1}_{i=0}{\beta^i}+\partial \theta^2\sum^{n-2}_{i=0}{\beta^i}+...+\partial \theta^n\sum^{0}_{i=0}{\beta^i}\bigg)=
$$
$$=(1-\beta)\bigg(\partial \theta^1\sum^n_{i=1}{\beta^{i-1}}+\partial \theta^2\sum^n_{i=2}{\beta^{i-2}}+...+\partial \theta^n\sum^n_{i=n}{\beta^{i-n}}\bigg)=
$$
$$=\sum^n_{s=1}{\bigg((1-\beta)\sum^n_{i=s}{\beta^{i-s}\bigg)\partial \theta^s}}=\sum^n_{s=1}{\gamma_s \theta^s}
$$

The above equality shows that taking $n$ steps using SGD with momentum produces the same state as the proposed optimization method. Additionally, both methods (SGD with momentum and the new method defined above) use the same matrices ${\theta_i}_{i=1}^n$ to reach that state (up to scalar multiplications). It implies similar expressive abilities, and therefore both processes are equivalent in that term.

Overall, SGD with momentum could be represented as the new variant of SGD we proposed. As explained earlier in this section, the new variant of SGD has an equivalent optimization process to a single layer. Therefore, the optimization process of SGD with momentum has an equivalent optimization process to a single fully-connected layer.

\section{Summary and open problems}
\label{conc}
We experimentally demonstrated how the derivatives of the weights $\partial \theta_l^t[j]$ (represented as vectors) are proportional for the same iteration and the same layer when training a deep fully-connected network with conventional optimizers. Provided with this experimental outcome, we concluded that a deep fully-connected network trained with conventional optimizers has an equivalent optimization process to a single fully-connected layer.

Our experiments are valid solely for fully-connected networks. Intuitively, the proportion between two vectors of weights is due to the mutual input layer observations during the training. Even though convolutional layers observe local regions individually (rather than the entire input simultaneously), there are still dependencies between close areas in the image and different filters concerning the same region. Therefore, deep linear convolutional neural networks probably do not have a convex optimization process; however, these models still might demonstrate interesting optimization constraints. Testing this hypothesis in future work might equip us with more knowledge regarding linear networks in particular and convolutional behavior in general.

\bibliography{main}

\appendix

\section{Additional experiments}
\label{appendix:additional_results_linear}
\begin{figure}[htp!]
     \includegraphics[width=1\textwidth]{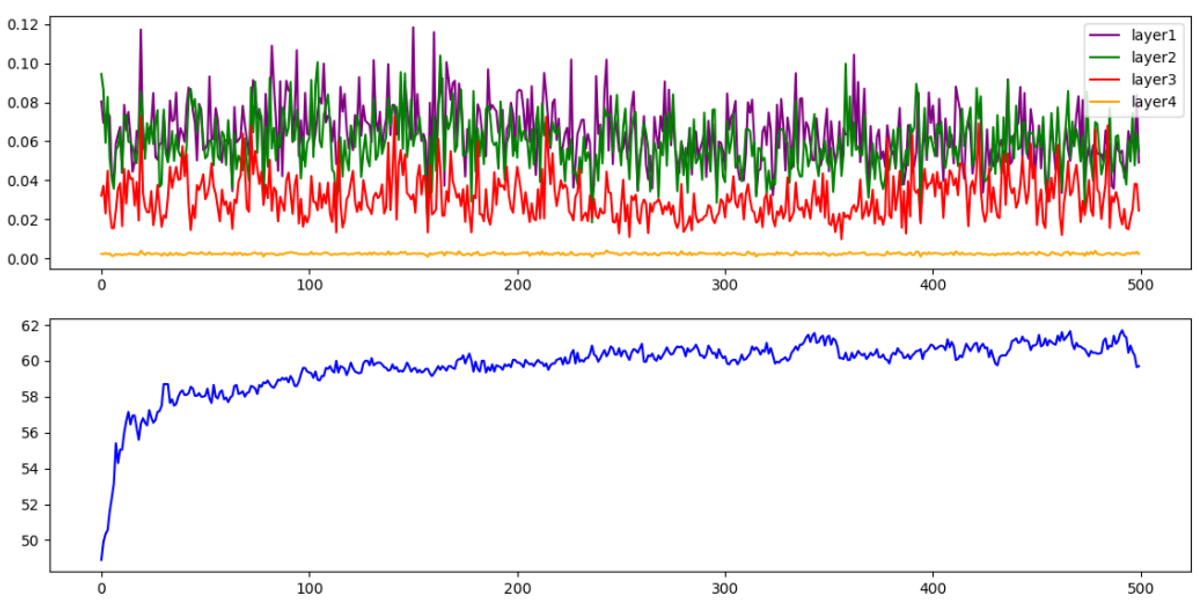}
     \caption{\emph{Cat} versus \emph{Dog} - a batch size of $128$, a learning rate of $1e-2$, and Architecture B trained for $500$ steps.}
     \label{fig:app_redundancy_3vs5_500}
\end{figure}

\begin{figure}[htp!]
     \includegraphics[width=1\textwidth]{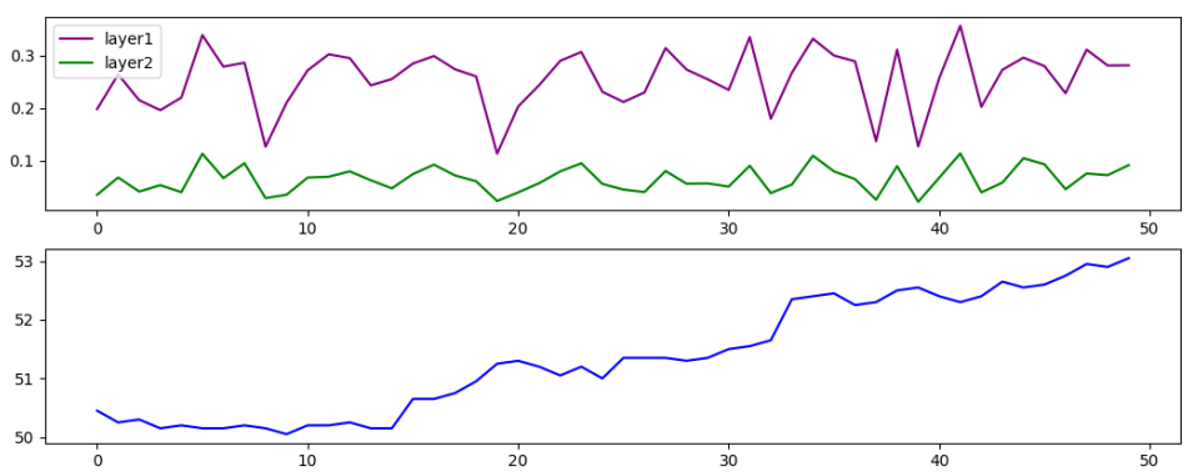}
     \caption{\emph{Ship} versus \emph{Truck} - a batch size of $256$, a learning rate of $1e-4$, and Architecture A trained for $50$ steps.}
     \label{fig:app_redundancy_8vs9_50}
\end{figure}

\begin{figure}[htp!]
     \includegraphics[width=1\textwidth]{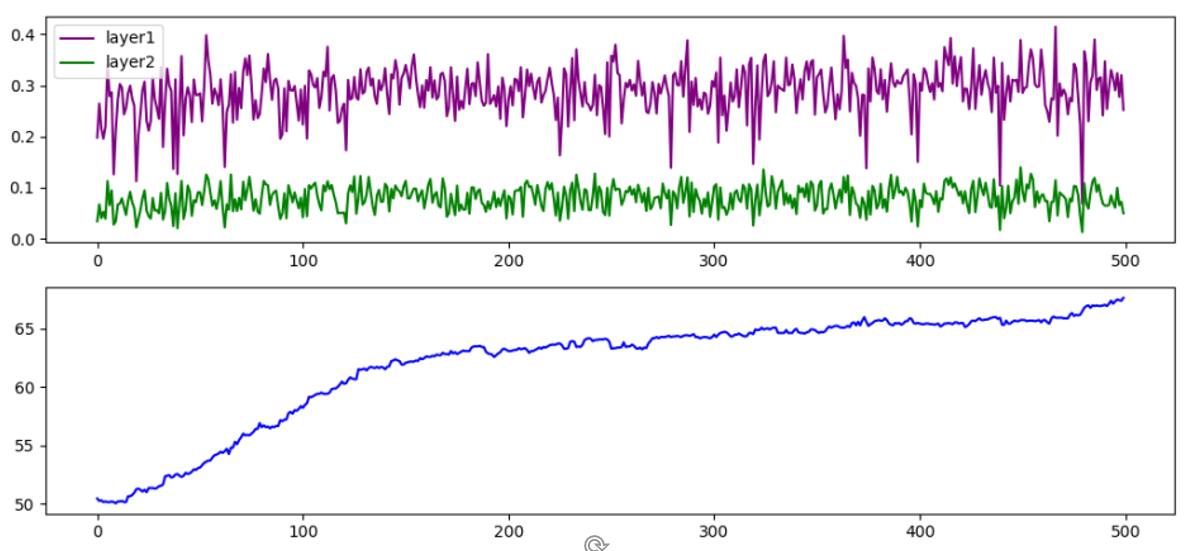}
     \caption{\emph{Ship} versus \emph{Truck} - a batch size of $256$, a learning rate of $1e-4$, and Architecture A trained for $500$ steps.}
     \label{fig:app_redundancy_8vs9_500}
\end{figure}

\begin{figure}[htp!]
     \includegraphics[width=1\textwidth]{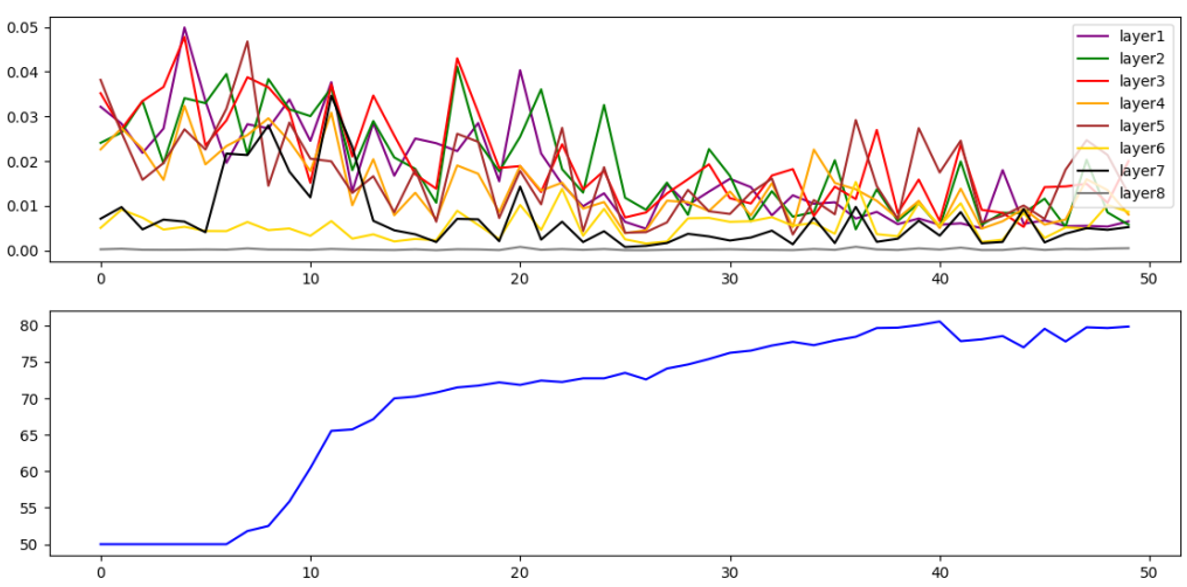}
     \caption{\emph{Airplane} versus \emph{Automobile} - a batch size of $64$, a learning rate of $1e-1$, and Architecture C trained for $50$ steps.}
     \label{fig:app_redundancy_0vs1_50}
\end{figure}

\begin{figure}[htp!]
     \includegraphics[width=1\textwidth]{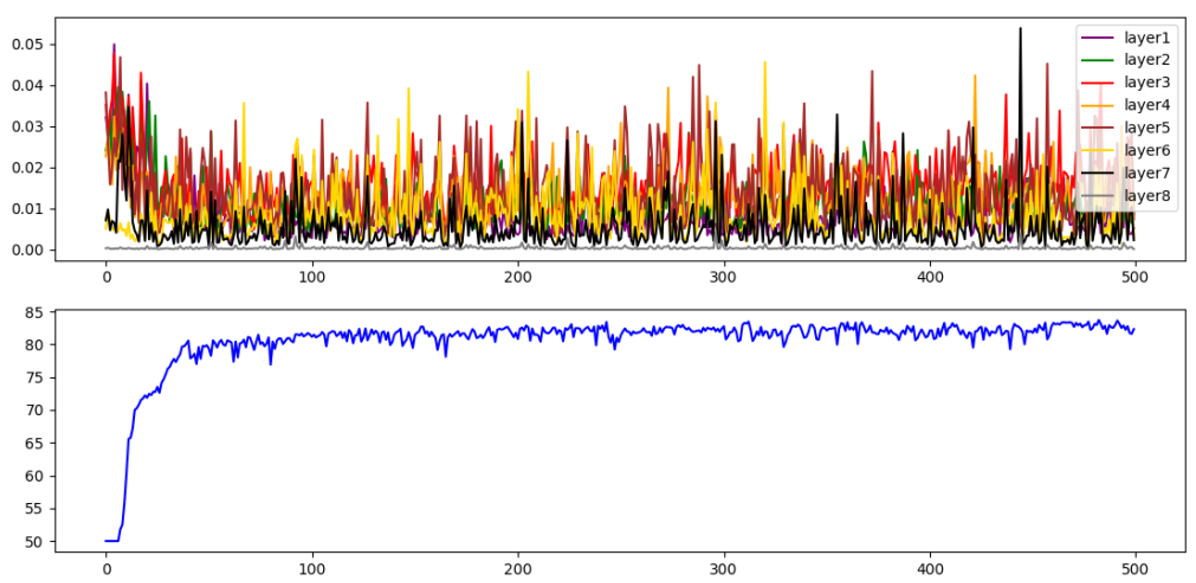}
     \caption{\emph{Airplane} versus \emph{Automobile} - a batch size of $64$, a learning rate of $1e-1$, and Architecture C trained for $500$ steps.}
     \label{fig:app_redundancy_0vs1_500}
\end{figure}

In Figures \ref{fig:app_redundancy_3vs5_500},\ref{fig:app_redundancy_8vs9_50},\ref{fig:app_redundancy_8vs9_500},\ref{fig:app_redundancy_0vs1_50},\ref{fig:app_redundancy_0vs1_500}, we can see the graph analysis of various cases with various learning rates, batch sizes, architectures (see Appendix \ref{appendix:architectures}), categories (of CIFAR10), and training resolutions.

\section{Architectures}
\label{appendix:architectures}

\begin{description}
\item[Architecture A:] Two linear layers.
\begin{itemize}
    \item \textbf{Fully-connected layer} [input: 3072, output: 128]
    \item \textbf{Fully-connected layer} [input: 128, output: 2]
\end{itemize}

\item[Architecture B:] Four linear layers.
\begin{itemize}
    \item \textbf{Fully-connected layer} [input: 3072, output: 2048]
    \item \textbf{Fully-connected layer} [input: 2048, output: 1024]
    \item \textbf{Fully-connected layer} [input: 1024, output: 128]
    \item \textbf{Fully-connected layer} [input: 128, output: 2]
\end{itemize}

\item[Architecture C:] Eight linear layers.
\begin{itemize}
    \item \textbf{Fully-connected layer} [input: 3072, output: 2500]
    \item \textbf{Fully-connected layer} [input: 2500, output: 1500]
    \item \textbf{Fully-connected layer} [input: 1500, output: 1024]
    \item \textbf{Fully-connected layer} [input: 1024, output: 512]
    \item \textbf{Fully-connected layer} [input: 512, output: 256]
    \item \textbf{Fully-connected layer} [input: 256, output: 64]
    \item \textbf{Fully-connected layer} [input: 64, output: 16]
    \item \textbf{Fully-connected layer} [input: 16, output: 2]
\end{itemize}
\end{description}

\end{document}